# Fuzzy thresholding in wavelet domain for speckle reduction in Synthetic Aperture Radar images

Mario Mastriani

*Abstract*—The application of wavelet transforms to Synthetic Aperture Radar (SAR) imagery has improved despeckling performance. To deduce the problem of filtering the multiplicative noise to the case of an additive noise, the wavelet decomposition is performed on the logarithm of the image gray levels. The detail coefficients produced by the bidimensional discrete wavelet transform (DWT-2D) needs to be thresholded to extract out the speckle in highest subbands. An initial threshold value is estimated according to the noise variance. In this paper, an additional fuzzy thresholding approach for automatic determination of the rate threshold level around the traditional wavelet noise thresholding (initial threshold) is applied, and used for the soft or hard-threshold performed on all the high frequency subimages. The filtered logarithmic image is then obtained by reconstruction from the thresholded coefficients. This process is applied a single time, and exclusively to the first level of decomposition. The exponential function of this reconstructed image gives the final filtered image. Experimental results on test images have demonstrated the effectiveness of this method compared to the most of methods in use at the moment.

## I. Introduction

SAR technology has resulted in marked improvements in the spatial resolution images when observing a ground scene from aircraft or satellites, and it can be used to estimate also features like the dampness of the soil, the thickness of a forest, or the roughness of the sea. Nevertheless, SAR images are contaminated by multiplicative noise, due to the coherence of the radar wavelength, labeled as speckle noise which results in an important reduction in the efficiency of target detection and classification algorithms.

Typical noise-smoothing methods are not well suited to preserving edge structures in speckled images. Classical operators are based on the local variance statistics [1]-[6]. On the other hand, and unfortunately, the thresholding technique has the following disadvantages:

1) it depends on the correct election of the type of thresholding, e.g., VisuShrink (soft-thresholding, hard-thresholding, and semi-soft-thresholding), SureShrink, OracleShrink, OracleThresh, BayesShrink, Thresholding Neural Network (TNN), etc. [2], [5]-[17],
2) it depends on the correct estimation of the threshold which is arguably the most important design parameter,
3) it doesn't have a fine adjustment of the threshold after their calculation,
4) it should be applied at each level of decomposition, needing several levels, and
5) the specific distributions of the signal and noise may not be well matched at different scales.

Therefore, a new method without these constraints will represent an upgrade. The method proposed here starts from a wavelet representation of the image. A few attempts were made at filtering of SAR images by wavelet, essentially filtering can be reduced hence to the case of an additive noise that is mastered in the framework of threshold method [8]-[10], [18]. The Bidimensional Discrete Wavelet Transform (DWT-2D) is outlined in Section II. In Section III, the method to reduce speckle noise by wavelet thresholding and the initial threshold estimation is discussed briefly. Section IV describes the additional fuzzy thresholding approach. In Section V, the speckle model. In Section VI, the experimental results using the proposed algorithm are presented. Finally, Section VII provides a conclusion of the paper.

## II. Bidimensional Discrete Wavelet Transform

The DWT-2D [19]-[28] corresponds to multiresolution approximation expressions. In practice, mutiresolution analysis is carried out using 4 channel filter banks composed of a low-pass and a high-pass filter and each filter bank is then sampled at a half rate (1/2 down sampling) of the previous frequency. By repeating this procedure, it is possible to obtain wavelet transform of any order. The down sampling procedure keeps the scaling parameter constant (equal to ½) throughout successive wavelet transforms so that is benefits for simple computer implementation. In the case of an image, the filtering is implemented in a separable way be filtering the lines and columns.

Note that [19] the DWT of an image consists of four frequency channels for each level of decomposition. For exam-

ple, for *i*-level of decomposition one has:

$CDD_{s,i}$ : Coefficients of Diagonal Detail, speckled,
$CVD_{s,i}$ : Coefficients of Vertical Detail, speckled,
$CHD_{s,i}$ : Coefficients of Horizontal Detail, speckled, and
$CA_i$ : Coefficients of Approximation.

The CA part at each scale is decomposed recursively, as illustrated in Fig. 1.

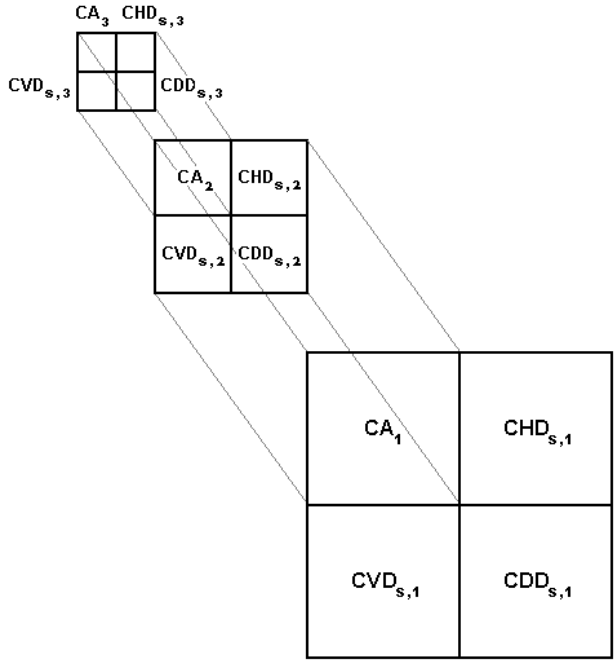

Fig. 1 Data preparation of the image. Recursive decomposition of CA parts.

To achieve space-scale adaptive noise reduction, one need to prepare the 1-D coefficient data stream which contains the space-scale information of 2-D images. This is somewhat similar to the "zigzag" arrangement of the DCT (Discrete Cosine Transform) coefficients in image coding applications [29].

| 64 | 63 | 15 | 45 | 3 | 9 | 33 | 39 |
|----|----|----|----|----|----|----|----|
| 62 | 61 | 30 | 60 | 6 | 12 | 36 | 42 |
| 14 | 44 | 13 | 43 | 18 | 24 | 48 | 54 |
| 29 | 59 | 28 | 58 | 21 | 27 | 51 | 57 |
| 2 | 8 | 32 | 38 | 1 | 7 | 31 | 37 |
| 5 | 11 | 35 | 41 | 4 | 10 | 34 | 40 |
| 17 | 23 | 47 | 53 | 16 | 22 | 46 | 52 |
| 20 | 26 | 50 | 56 | 19 | 25 | 49 | 55 |

Fig. 2 Data preparation of the image. Spatial order of 2-D coefficients.

In this data preparation step, the DWT-2D coefficients are rearranged as a 1-D coefficient series in spatial order so that the adjacent samples represent the same local areas in the original image.

An example of the rearrangement of an 8-by-8 transformed image is shown in Fig. 2, which will be referred to as a 1-D space-scale data stream.

Each number in Fig. 2 represents the spatial order of the 2-D coefficient at that position corresponding to Fig. 1.

### III. WAVELET NOISE THRESHOLDING

The wavelet coefficients calculated by a wavelet transform represent change in the image at a particular resolution. By looking at the image in various resolutions it should be possible to filter out noise. At least in theory. However, the definition of noise is a difficult one. In fact, "one person's noise is another's signal". In part this depends on the resolution one is looking at. One algorithm to remove Gaussian white noise is summarized by D.L. Donoho and I.M. Johnstone [9], and synthesized in Fig. 3.

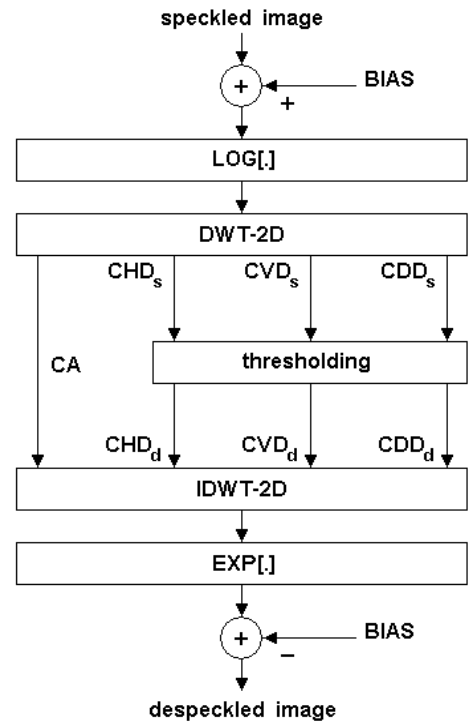

Fig. 3 Thresholding Techniques

where:
*+BIAS* is applied to the whole image, and it is for avoiding;
$LOG[0] = -\infty$, when a pixel have a luminance value = 0;
*LOG[•]* is the natural logarithm of the elements of [•];
*DWT-2D* is the Bidimensional Discrete Wavelet Transform;
*IDWT-2D* is the inverse of DWT-2D

*EXP[•]* is the exponential of the elements of [•]; and *-BIAS* restores to +BIAS.

The algorithm is:
1) Calculate a wavelet transform and order the coefficients by increasing frequency. This will result in an array containing the image average plus a set of coefficients of length 1, 2, 4, 8, etc. The noise threshold will be calculated on the highest frequency coefficient spectrum (this is the largest spectrum).

2) Calculate the *median absolute deviation* on the largest coefficient spectrum. The median is calculated from the absolute value of the coefficients. The equation for the median absolute deviation is shown below:

$$\delta_{mad} = \frac{median(|CD_{s,i}|)}{0.6745} \quad (1)$$

where $CD_{s,i}$ may be $CHD_{s,i}$, $CVD_{s,i}$, or $CDD_{s,i}$ for *i*-level of decomposition. The factor 0.6745 in the denominator rescales the numerator so that $\delta_{mad}$ is also a suitable estimator for the standard deviation for Gaussian white noise [9]. In fact, the controller will include any type of noise.

3) For calculating the noise threshold λ, a modified version of the equation that has been discussed in papers by D.L. Donoho and I.M. Johnstone is used. The equation is:

$$\lambda = \delta_{mad}\sqrt{2log[N]} \quad (2)$$

where N is the number of pixels in the subimage, i.e., CVD, CHD or CDD.

4) Apply a thresholding algorithm to the coefficients. There are two popular versions:

4.1. Hard thresholding. Hard thresholding sets any coefficient less than or equal to the threshold to zero, see Fig. 4(a).

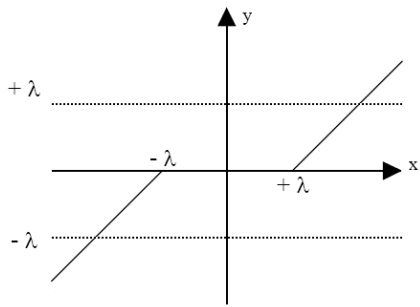

Fig. 4(a) Soft-Thresholfing

where *x* may be $CHD_{s,i}$, $CVD_{s,i}$, or $CDD_{s,i}$, *y* may be $CDD_{d,i}$: Coefficients of Diagonal Detail, de-speckled, $CVD_{d,i}$: Coefficients of Vertical Detail, de-speckled, or $CHD_{d,i}$: Coefficients of Horizontal Detail, de-speckled.for *i*-level of decomposition.

The respective code is:

```
for row = 1:N^(1/2)
  for column = 1:N^(1/2)
    if |CDs,i[row][column]| <= λ,
      CDs,i[row][column] = 0.0;
    end
  end
end
```

4.2. Soft thresholding. Soft thresholding sets any coefficient less than or equal to the threshold to zero, see Fig. 4(b). The threshold is subtracted from any coefficient that is greater than the threshold. This moves the image coefficients toward zero.

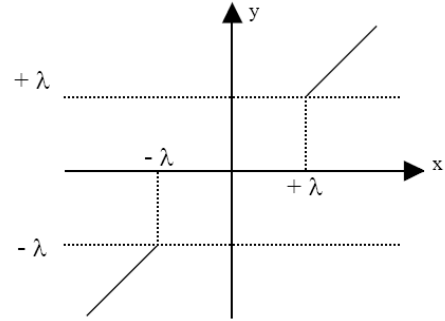

Fig. 4(b) Hard-Thresholfing

The respective code is:

```
for row = 1:N^(1/2)
  for column = 1:N^(1/2)
    if |CDs,i[row][column]| <= λ,
      CDs,i[row][column] = 0.0;
    else
      CDs,i[row][column] = CDs,i[row][column] - λ;
    end
  end
end
```

IV. FUZZY THRESHOLDING IN WAVELET DOMAIN (FUZZYTHRESH)

Although, the transform can be any DWT-2D [19], however, for a specific class of image, the appropriate DWT-2D may be selected to concentrate image energy versus noise. By thresholding [8]-[10], the image energy may be kept while the noise is suppressed. The IDWT-2D is employed to recover the signal from the noise-reduced coefficients in transform domain. Since many images have some regularities and DWT-2D is a very good tool to efficiently represent such characteristics of the image. On the other hands, DWT-2D is often a suitable linear orthogonal transform. This will be a meaningful exploration that will be pursued in the future.

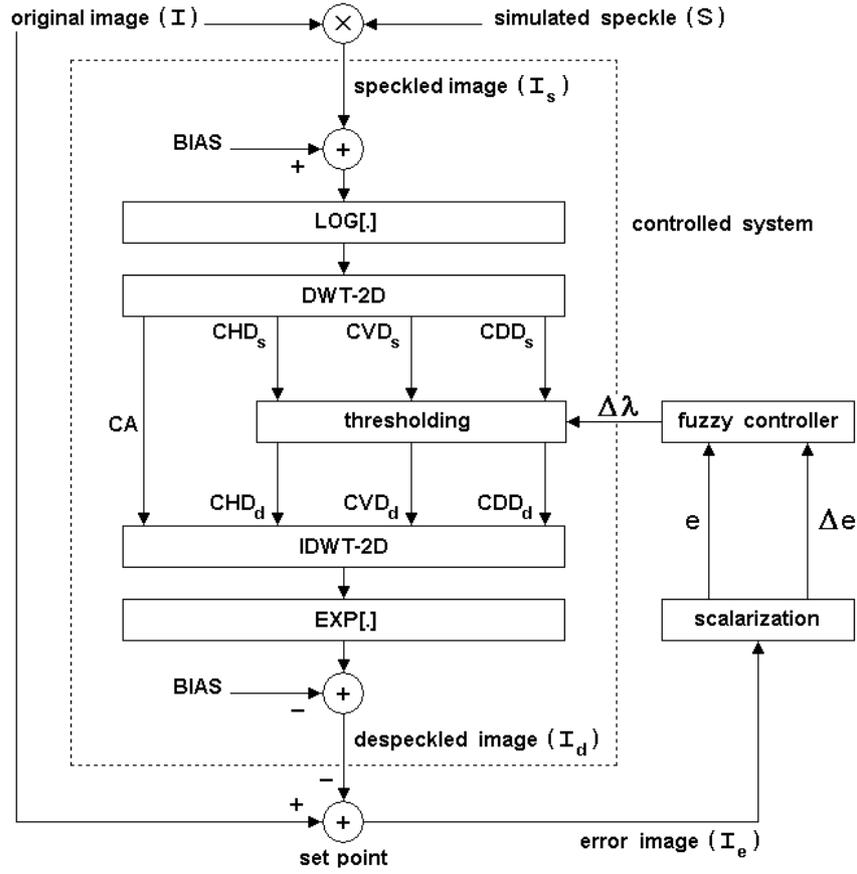

Fig. 5 FuzzyThresh with detail of controlled system inside of dot line.

Fuzzy controller [30]-[34] is the elected tool to improve the thresholding process, which is directly applied to Coefficients of Detail (CHD, CVD and CDD) of DWT-2D. The Fuzzy Thresholding structure proposed for our experiment can be illustrated as in Fig. 5, with:

$$I_e = I - I_d \qquad (3)$$

where $I_e$ is the error image,
  $I$ is the original image (without speckle), and
  $I_d$ is the despeckled image.

On the other hand, two scalar variables $e$ and $\Delta e$ are obtained from the *scalarization* process of Fig. 6, where: $|\bullet|$ means modulus of $(\bullet)$; $max(\bullet)$ means maximum of $(\bullet)$; $e$ means *scalar error; and* $\Delta e$ means *Change in scalar error.* Initial values of $e$, $\Delta e$ and $\Delta \lambda$ are zero.

Returning to the Fig. 5, the output of the *fuzzy controller* is the *change in threshold* $\Delta \lambda$, which will used as shown in Fig. 7 for Soft and Hard-Thresholding respectively, vely, where $\lambda$ was calculated in (2). Besides, the fuzzy controller in detail is shown in Fig. 8 and, the controlled system was shown in detail in Fig. 5.

Fuzzy control [30], [31] is a control method based on the theory of fuzzy sets and fuzzy logic [32], [33]. Just as fuzzy logic can be described simply as "computing with words rather than numbers", fuzzy control can be described simply as ''control with sentences rather than equations''. A fuzzy controller can include empirical rules, and that is especially useful in operator controlled plants.

Take for instance a typical fuzzy controller

1) *If error is Negative Small (NS) and change in error is NS then output is NS*

2) *If error is NS and change in error is Positive Small (PS) then output is Approximately Zero (AZ),* etc.

The collection of rules is called a *rule base*. The rules are in the familiar if-then format, and formally the if-side is called the *condition* and the then-side is called the *conclusion* (more often, perhaps, the pair is called *antecedent-consequent* or *premise-conclusion*). The computer is able to execute the rules and compute a control signal depending on the measured inputs *error* and *change in error*. In Table I the linguistic variables for the control rules of Table II can be seen.

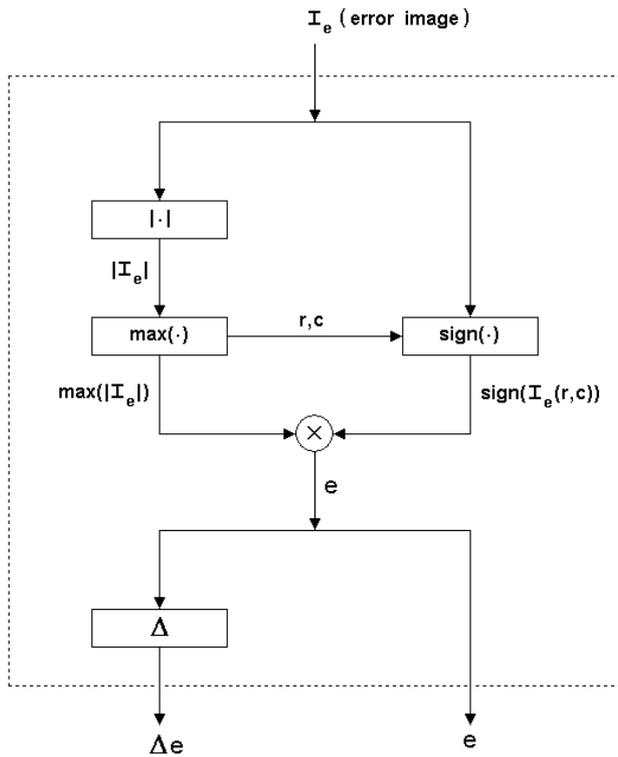

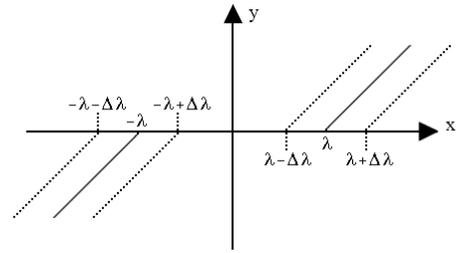

Fig. 7(a) Fuzzy Control in Soft-Thresholding

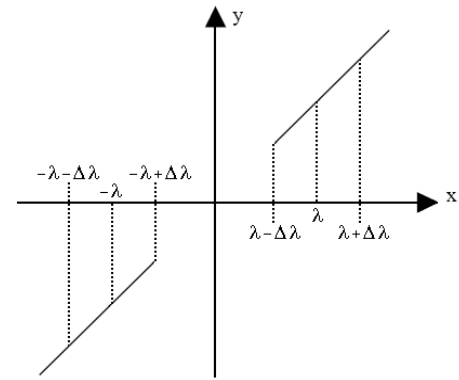

Fig. 6 Scalarization Process

Fig. 7(b) Fuzzy Control in Hard-Thresholding

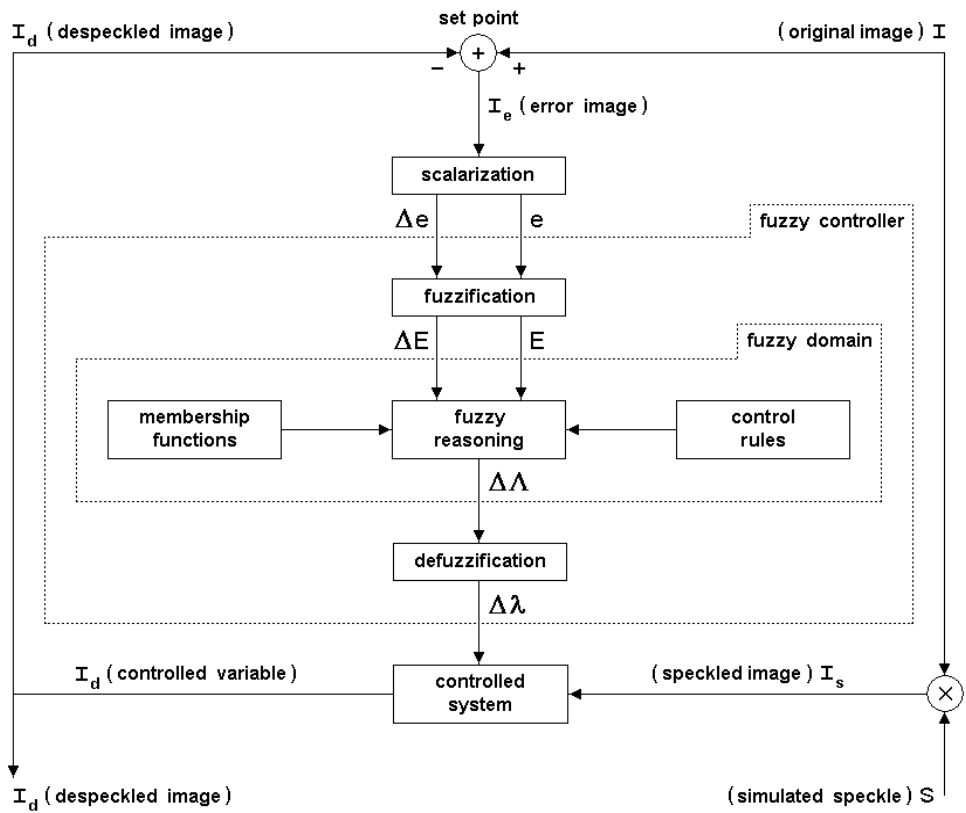

Fig. 8 Fuzzy Control with detail of fuzzy controller inside of outer dot line.

TABLE I
QUANTIZATION LEVEL AND VALUES OF PB, PS, AZ, NS, AND NB

| System Variables | Signal Values | | | | | | | | |
|---|---|---|---|---|---|---|---|---|---|
| Error, **e** | -1 | -0.75 | -0.5 | -0.25 | 0 | 0.25 | 0.5 | 0.75 | 1 |
| Change in error, $\Delta$**e** | -1 | -0.75 | -0.5 | -0.25 | 0 | 0.25 | 0.5 | 0.75 | 1 |
| Control, $\Delta\lambda$ | -1 | -0.75 | -0.5 | -0.25 | 0 | 0.25 | 0.5 | 0.75 | 1 |
| Quantification level | -4 | -3 | -2 | -1 | 0 | 1 | 2 | 3 | 4 |

| Linguistic variables | Membership Values | | | | | | | | |
|---|---|---|---|---|---|---|---|---|---|
| Positive Big, PB | 0 | 0 | 0 | 0 | 0 | 0 | 0 | 0.5 | 1 |
| Positive Small, PS | 0 | 0 | 0 | 0 | 0 | 0.5 | 1 | 0.5 | 0 |
| Approximately Zero, AZ | 0 | 0 | 0 | 0.5 | 1 | 0.5 | 0 | 0 | 0 |
| Negative Small, NS | 0 | 0.5 | 1 | 0.5 | 0 | 0 | 0 | 0 | 0 |
| Negative Big, NB | 1 | 0.5 | 0 | 0 | 0 | 0 | 0 | 0 | 0 |

TABLE II
CONTROL RULES FOR THE LINGUISTIC VARIABLES IN TABLE II

| $\Delta$e \ e | NB | NS | AZ | PS | PB |
|---|---|---|---|---|---|
| NB | NB | NS | NS | AZ | AZ |
| NS | NB | NS | AZ | AZ | PS |
| AZ | NS | NS | AZ | PS | PS |
| PS | NS | AZ | AZ | PS | PB |
| PB | AZ | AZ | PS | PS | PB |

$\Delta\lambda$

The principle behind the technique is that imprecise data can be classified into sets having fuzzy rather than sharp boundaries, which can be manipulated to provide a framework for approximate reasoning in the face of imprecise and uncertain information.

A fuzzy expert system functions in four steps. The first is *fuzzification*, during which the membership functions defined on the input variables are applied to their actual values, to determine the degree of truth for each rule premise. The membership functions of the fuzzy sets and the fuzzy control rules have a big effect on control performance. Fig. 9 shows the membership functions of the fuzzy sets for the position errors, velocity errors, and control inputs.

Next under *inference*, the truth value for the premise of each rule is computed, and applied to the conclusion part of each rule. This results in one fuzzy set to be assigned to each output variable for each rule.

In *composition*, all of the fuzzy sets assigned to each output variable are combined together to form a single fuzzy set for each output variable. Finally comes *defuzzification*, which converts the fuzzy output set to a crisp (nonfuzzy) number. Among many defuzzification strategies, the center average method is adopted [34].

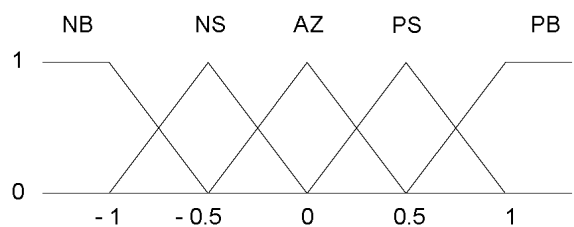

Fig. 9 Membership functions for Error **e**, Change in Error $\Delta$**e**, and Change in Threshold $\Delta\lambda$.

A fuzzy controller may then be designed using a fuzzy expert system to perform fuzzy logic operations on fuzzy sets representing linguistic variables in a qualitative set of control rules, see Fig. 8.

As a simple metaphor of fuzzy control in practice, consider the experience of balancing a stick vertically on the palm of ones hand. The equations of motion for the stick (a pendulum at its unstable fixed point) are well-known, but these equations should not be integrated in order to balance the stick. Rather, one stare at the top of the stick and carry out a type of fuzzy control to keep the stick in the air: one move our hand slowly when the stick leans by a small angle, and fast when it leans by a larger angle. Our ability to balance the stick despite the imprecision of our knowledge of the system is at the heart of fuzzy control. In definitive,

- input of the fuzzy controller are error **e** and change in error $\Delta e$, i.e., one has 2-dimensional rule base, see Fig. 5, 6 and 8,

- output of the fuzzy controller can be either control signal $\lambda$ (threshold) or change in control signal $\Delta\lambda$ (change in threshold), because one has two types of fuzzy controllers:

1. if the output is control signal (threshold), the fuzzy controller is said to be of Proportional-Derivative-type (PD-type), i.e., PD-controller

$$\lambda = k_P (e + T_D \Delta e) \quad (4)$$

where $k_P$ is the controller gain and $T_D$ is the *derivative time*.

2. a fuzzy controller of Proportional-Integral-type (PI-type) yields change in control signal (change in threshold), i.e., PI-controller

$$\Delta\lambda = k_P (\Delta e + e / T_I) \quad (5)$$

where $T_I$ is the *integral time*. For experimental reasons one choose the last one.

- considering measurement as input, then, the controller can make use of the operating point. If inputs are $e$, $\Delta e$, and $\Delta\lambda$, the rule base is 3-dimen-sional, e.g., 5 membership functions for each input, then, one has $5^3$ rules.

- the fuzzy block calculates parameters ( $k_P$ and $T_I$ ) for the PI-controller (fuzzy gain scheduling).

- rule base in a table:

  1. a 2-dimensional rule base can be presented as a table, see Table II

  2. sign rules:
     2.1. $e$ is positive when measurement is smaller than set point, see Fig. 5 and 8
     2.2. $\Delta e$ is positive when measurement decreases (and set point is constant)

  3. the Table II is a typical PI-type fuzzy controller with a loose tuning (wide band of zeros in the diagonal) and with a typical output surface, see Fig. 10

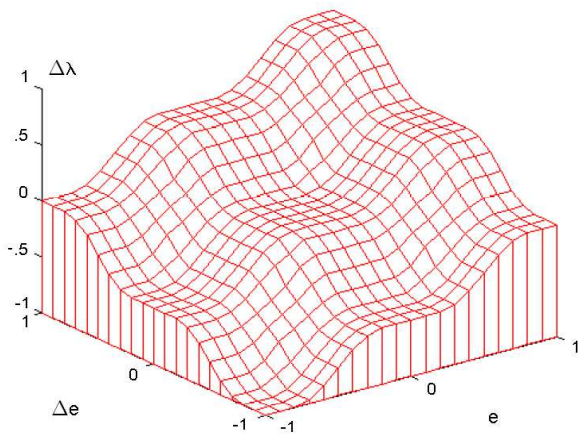

Fig. 10 Output surface

- tuning a fuzzy controller. The control can be improved by

  1. a complete rule base, if there are holes in the rule base, the implementation of the controller decides what happens at an "empty" rule: the output either remains constant or, at worst, falls to zero

  2. orthogonal term sets of the inputs, with appropriate reasoning method, S-norm and T-norm [30]-[34] also membership functions of the fuzzy relations become orthogonal, and the divider of bisector defuzzification can be omitted

  3. support of the input membership functions covers the whole universe of discourse ("holes" in support imply division by zero in bisector defuzzification)

- tuning methods

  1. setting membership functions
     1.1. knowledge of input ranges is needed, see Table I
     1.2. membership functions can be more dense where accurate control is needed

  2. tuning rule base
     2.1. process expertise is needed
     2.2. there are also systematic methods utilizing for example PI-controller parameters

  3. as in any controller tuning, practical methods include
     3.1. trial a error
     3.2. common sense
     3.3. experience

- tuning tools

  1. response (step response), as Fig. 11

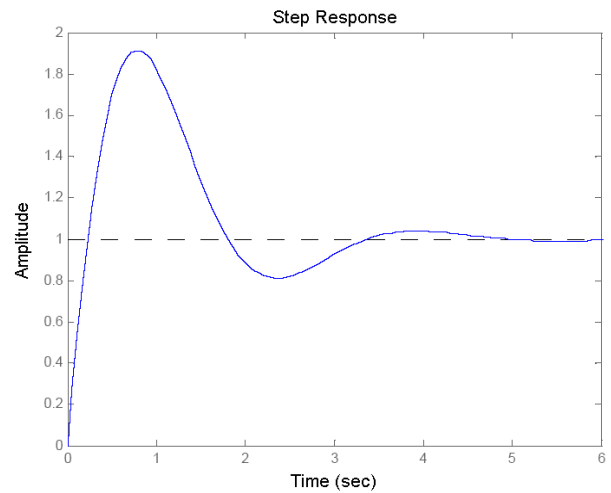

  2. trajectory, as Fig. 12

  3. rule base + trajectory, as Table III

  4. output surface + trajectory, show which rules contribute to the response, as Fig. 13

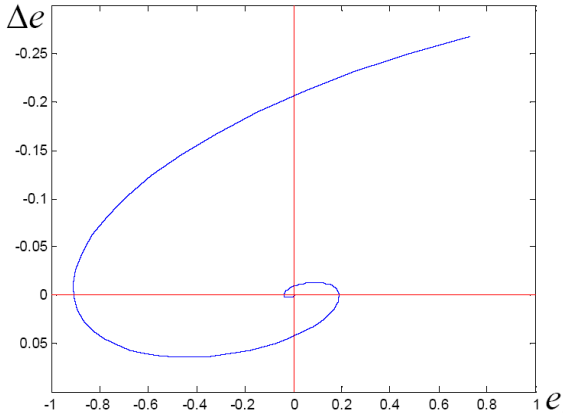

Fig. 12 Trajectory

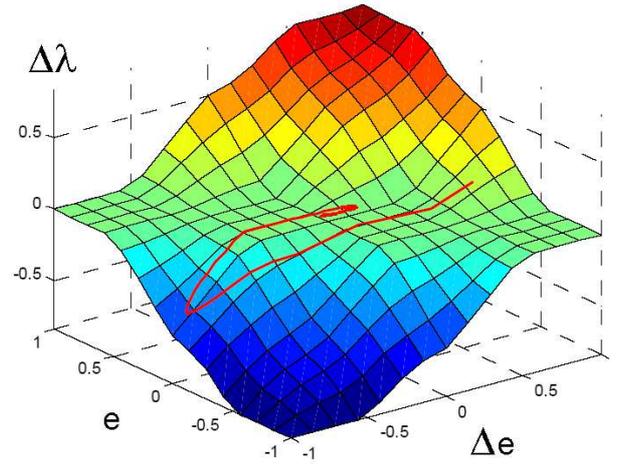

Fig. 13 Output surface + trajectory

TABLE III
RULE BASE + TRAJECTORY

| NB | NS | NS | AZ | AZ |
|----|----|----|----|----|
| NB | NS | AZ | AZ | PS |
| NS | NS | AZ | PS | PS |
| NS | AZ | AZ | PS | PB |
| AZ | AZ | PS | PS | PB |

Finally, the Complete Fuzzy Thresholding (as an unified algorithm in a pseudo-MATLAB code) to improve the noise thresholding can be represented as show in Algorithm I.

The simplicity of this algorithm allows an easy and cheap implementation on a chip [31], [32], for the fuzzy thresholding regarding its competitors.

Finally, and summing up, first synthetic speckle is applied on an image without speckle (synthetic speckle with the same statistical distribution of the real speckle) and once concluded the training (that is to say, calculated the final threshold) the FuzzyThresh is applied to ERS-2 images with real speckle [14].

ALGORITHM I

```
1.  % calculation of the noise threshold λ
      a.  Is = I * S;
      b.  % + BIAS
      c.  % LOG[.]
      d.  % DWT-2D
      e.  % lambda from Section III              % λ from Equation (2)
2.  me > epsilon > 0;                            % |e_initial| > ε (arbitrary) > 0;
3.  eh = me;                                     % historical error
4.  while(me>epsilon)
      a.  Ie = I - Id;                           % Equation (3)
      b.  % scalarization algorithm
            i.   mIe = abs(Ie);                  % | Ie |
            ii.  [MAX,row,column] = max(mIe);
            iii. SIG = sign(Ie(row,column));
            iv.  e = SIG * MAX;
            v.   De = e-eh;                      % change in error Δe
      c.  % fuzzy controller
            i.   Dlambda = fuzzy_controller(e,De); % change in threshold Δλ
      d.  % adjust of the noise threshold
            i.   lambda = lambda + Dlambda;      % λ = λ + Δλ
      e.  % update of Id
            i.   % thresholding of CHDs, CVDs, and CDDs
            ii.  % IDWT-2D
            iii. % EXP[.]
            iv.  % -BIAS
      f.  me = abs(e);                           % | e |
      g.  eh = e;                                % historical error
end
```

## V. SPECKEL MODEL

Speckle noise in SAR images is usually modelled as a purely multiplicative noise process of the form

$$I_s(r,c) = I(r,c) \, S(r,c) \qquad (6)$$

The true radiometric values of the image are represented by $I$, and the values measured by the radar instrument are represented by $I_s$. The speckle noise is represented by $S$. The parameters $r$ and $c$ means row and columns of the respective pixel of the image.

For single-look SAR images, $S$ is Rayleigh distributed (for amplitude images) or negative exponentially distributed (for intensity images) with a mean of 1. For multi-look SAR images with independent looks, $S$ has a gamma distribution with a mean of 1. Further details on this noise model are given in [35].

## VI. STATISTICAL MEASUREMENT

In this work, the assessment parameters that are used to evaluate the performance of speckle reduction are Noise Variance, Mean Square Difference, Noise Mean Value, Noise Standard Deviation, Equivalent Number of Looks, Deflection Ratio, and Pratt's figure of Merit [2], [36].

### A. Noise Mean Value (NMV), Noise Variance (NV), and Noise Standard Deviation (NSD)

$NV$ determines the contents of the speckle in the image. A lower variance gives a "cleaner" image as more speckle is reduced, although, it not necessarily depends on the intensity. The formulas for the $NMV$, $NV$ and $NSD$ calculation are

$$NMV = \frac{\sum_{r,c} I_d(r,c)}{R*C} \qquad (7.1)$$

$$NV = \frac{\sum_{r,c} (I_d(r,c) - NMV)^2}{R*C} \qquad (7.2)$$

$$NSD = \sqrt{NV} \qquad (7.3)$$

where $R$-by-$C$ pixels is the size of the despeckled image ($I_d$).

On the other hand, the estimated noise variance is used to determine the amount of smoothing needed for each case for all filters.

### B. Mean Square Difference (MSD)

$MSD$ indicates average difference of the pixels throughout the image where $I_d$ is the denoised image, and $I_s$ is the original image (with speckle), see Fig. 16. A lower $MSD$ indicates a smaller difference between the original (with speckle) and denoised image. This means that there is a significant filter performance. Nevertheless, it is nece-ssary to be very careful with the edges. The formula for the $MSD$ calculation is

$$MSD = \frac{\sum_{r,c} (I_s(r,c) - I_d(r,c))^2}{R*C} \qquad (8)$$

### C. Equivalent Numbers of Looks (ENL)

Another good approach of estimating the speckle noise level in a SAR image is to measure the $ENL$ over a uniform image region [15]. A larger value of $ENL$ usually corres-ponds to a better quantitative performance. The value of $ENL$ also depends on the size of the tested region, theoretically a larger region will produces a higher $ENL$ value than over a smaller region but it also tradeoff the accuracy of the readings. Due to the difficulty in identifying uniform areas in the image, to divide the image into smaller areas of 25x25 pixels is proposed, obtain the $ENL$ for each of these smaller areas and finally take the average of these $ENL$ values. The formula for the $ENL$ calculation is

$$ENL = \frac{NMV^2}{NSD^2} \qquad (9)$$

The significance of obtaining both $MSD$ and $ENL$ measurements in this work is to analyze the performance of the filter on the overall region as well as in smaller uniform regions.

### D. Deflection Ratio (DR)

A third performance estimator that is used in this work is the $DR$ proposed by H. Guo et al (1994), [15]. The formula for the deflection calculation is

$$DR = \frac{1}{R*C} \sum_{r,c} \left( \frac{I_d(r,c) - NMV}{NSD} \right) \qquad (10)$$

The ratio $DR$ should be higher at pixels with stronger reflector points and lower elsewhere. In H. Guo et al's paper, this ratio is used to measure the performance between different wavelet shrinkage techniques on the diagonal subband only. Instead the ratio approach to all subbands is applied [36].

### E. Pratt's Figure of Merit (FOM)

To compare edge preservation performances of different speckle reduction schemes, the Pratt's figure of merit is adopted [4] defined by

$$FOM = \frac{1}{max\{\hat{N}, N_{ideal}\}} \sum_{i=1}^{\hat{N}} \frac{1}{1+d_i^2 \alpha} \qquad (11)$$

Where $\hat{N}$ and $N_{ideal}$ are the number of detected and ideal edge pixels, respectively, $d_i$ is the Euclidean distance between the $i$th detected edge pixel and the nearest ideal edge pixel, and $\alpha$ is a constant typically set to 1/9. *FOM* ranges between 0 and 1, with unity for ideal edge detection.

## VII. RESULTS

The simulations demonstrate that the FuzzyThresh architecture improves the speckle reduction performance to the maximum, for SAR image.

Here, I present a set of experimental results using one ERS SAR Precision Image (PRI) standard of Buenos Aires area. Such image was converted to bitmap file format [29] for a better treatment.

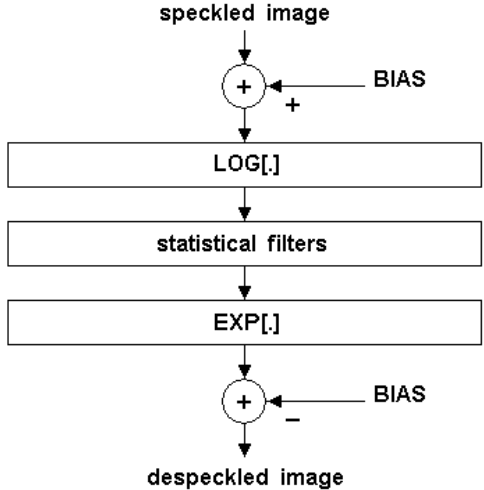

Fig. 14(a) Homomorphic reduction scheme for statistical filters.

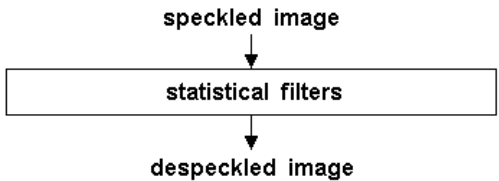

Fig. 14(b) Simple reduction scheme for statistical filters.

For the statistical filters: Median, Wiener [37]-[48], Directional Smoothing (DS) [4], and Enhanced DS (EDS) [36], a homomorphic speckle reduction scheme based on Fig. 14(a) is used, with 3-by-3 kernel windows; while for the rest: Lee, Kuan, Gamma-Map, Enhanced Lee, Frost, Enhanced Frost, [1]-[3], [37]-[48], a simple speckle reduction scheme based on Fig. 14(b) is used, because these filters were originally defined intrinsically multiplicative noise and a logarithm transformation is not only theoretically inco-rrect, but it can give rise to a complete misinterpretation of the final results. Besides, for Lee, Enhanced Lee, Kuan, Gamma-Map, Frost and Enhanced Frost filters the damping factor is set to different values depending on 3-look ampli-tude data, see [1]-[3], [37]-[48], with 5-by-5 and 7-by-7 kernel windows.

Fig. 15 shows a noisy image used in the experiment from remote sensing satellite ERS-2, with a 540-by-553 (pixels) by 65536 (gray levels) bitmap matrix. It will be obtained using a chirp scaling algorithm [49], [50].

Table IV summarizes the assessment parameters vs. 19 filters for Fig. 15, where En-Lee means Enhanced Lee Filter, En-Frost means Enhanced Frost Filter, ST means Soft-Thresholding, HT means Hard-Thresholding and SST means Semi-Soft-Thresholding. The quantitative results of Table IV show that the FuzzyThresh can eliminate speckle without distorting useful image information and without destroying the important image edges.

Fig. 16 shows the filtered images in the experiment, processed by using VisuShrink (HT) [14] with Daubechies 15 wavelet basis and 1 level of decomposition (improvements were not noticed with other basis of wavelets) [5], [6], [16]-[18], [20]-[28], [51]-[54], BayesShrink, OracleShrink, SUREShrink, and FuzzyThresh respectively. Such filters was applied to complete image, however, only a selected 242-by-242 pixels windows is showed for space considerations.

On the other hand, Fig. 16 summarizes the edge preservation performance of the new FuzzyThresh vs. the rest of the filters with a considerably smaller computational complexity. As shown in Fig. 16, the numerical results are further supported by qualitative examination.

NMV and NSD are computed and compared tover three different homogeneous regions in our SAR image, before and after filtering, for all filters. The FuzzyThresh has obtained the best mean preservation and variance reduction, as shown in Table IV. Since a successful speckle reducing filter will not significantly alter the mean intensity within a homogeneous region, FuzzyThresh demonstrated to be the best in this sense too.

In the experiment, the FuzzyThresh outperformed the conventional and no conventional speckle reducing filters in terms of edge preservation measured by Pratt figure of merit [4]. In nearly every case in every homogeneous region, the FuzzyThresh produced the lowest standard deviation and were able to preserve the mean value of the region.

For TNN [14] the empirical function parameter value $\lambda$ = 0.01, and the used activation function was the hyperbolic tangent for input and hidden layer,

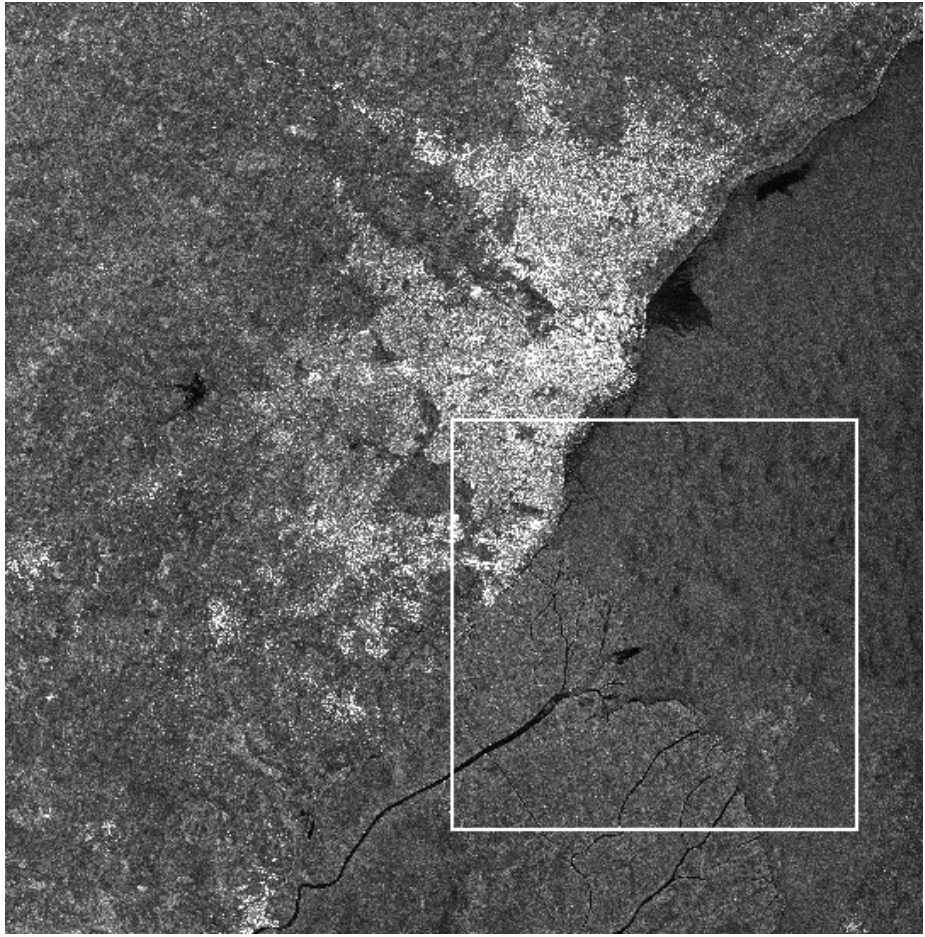

Fig. 15: ERS-2 image for the experiment of Table IV. The white square represents the perimeter of the selected windows (242-by-242 pixels) used in Fig. 16.

TABLE IV
ASSESSMENT PARAMETERS VS. FILTERS FOR FIG. 15

| Filter | Assessment Parameters | | | | | | |
|---|---|---|---|---|---|---|---|
| | NV | MSD | NMV | NSD | ENL | DR | FOM |
| Original noisy image | 2.55379e+004 | - | 276.82 | 159.80 | 3.0006 | 0.0004 | 0.3027 |
| En-Frost | 1.45335e+004 | 931.0141 | 261.11 | 120.55 | 4.6911 | 0.0014 | 0.4213 |
| En-Lee | 1.48322e+004 | 930.8242 | 262.98 | 121.78 | 4.6627 | 0.0014 | 0.4112 |
| Frost | 1.48638e+004 | 649.7037 | 255.14 | 121.91 | 4.3795 | 0.0014 | 0.4213 |
| Lee | 1.28158e+004 | 939.4810 | 251.44 | 113.20 | 4.9331 | 0.0014 | 0.4228 |
| Gamma-MAP | 1.41052e+004 | 932.9512 | 248.66 | 118.76 | 4.3836 | 0.0016 | 0.4312 |
| Kuan | 1.54233e+004 | 406.7207 | 252.34 | 124.19 | 4.1285 | 0.0013 | 0.4217 |
| Median | 1.57190e+004 | 931.3837 | 271.33 | 125.37 | 4.6835 | 0.0014 | 0.4004 |
| Wiener | 1.81585e+004 | 762.3237 | 277.45 | 134.75 | 4.2255 | 0.0013 | 0.4423 |
| DS | 1.42273e+004 | 255.1525 | 278.90 | 119.27 | 5.4673 | 0.0031 | 0.4572 |
| EDS | 1.31279e+004 | 225.7491 | 277.88 | 114.57 | 5.8819 | 0.0032 | 0.4573 |
| VisuShrink (ST) | 1.34976e+004 | 888.5013 | 284.90 | 116.17 | 6.0135 | 0.0031 | 0.4519 |
| VisuShrink (HT) | 1.30047e+004 | 885.6094 | 285.05 | 114.03 | 6.2480 | 0.0033 | 0.4522 |
| VisuShrink (SST) | 1.24232e+004 | 878.6534 | 278.99 | 111.45 | 6.2653 | 0.0032 | 0.4521 |
| SureShrink | 1.13540e+004 | 891.6578 | 280.15 | 106.55 | 6.9124 | 0.0029 | 0.4520 |
| OracleShrink | 1.36980e+004 | 888.3241 | 290.15 | 117.03 | 6.1459 | 0.0029 | 0.4576 |
| BayesShrink | 1.23092e+004 | 859.3426 | 291.35 | 110.94 | 6.8960 | 0.0033 | 0.4581 |
| OracleThresh | 1.24057e+004 | 232.7623 | 291.22 | 111.38 | 6.8363 | 0.0036 | 0.4581 |
| TNN | 1.17488e+004 | 185.9812 | 282.43 | 108.39 | 6.7893 | 0.0039 | 0.4588 |
| FuzzyThresh | 9.80759e+003 | 182.1244 | 276.81 | 99.03 | 7.8127 | 0.0041 | 0.4590 |

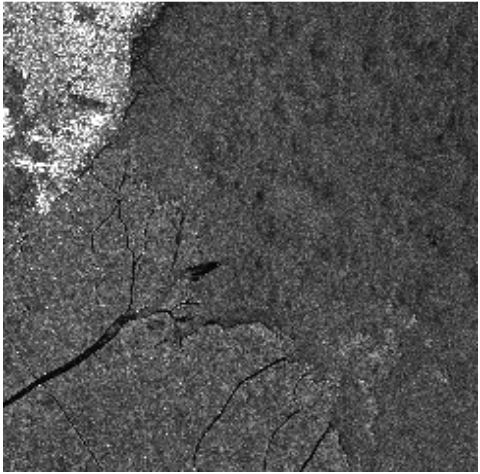

(a) original

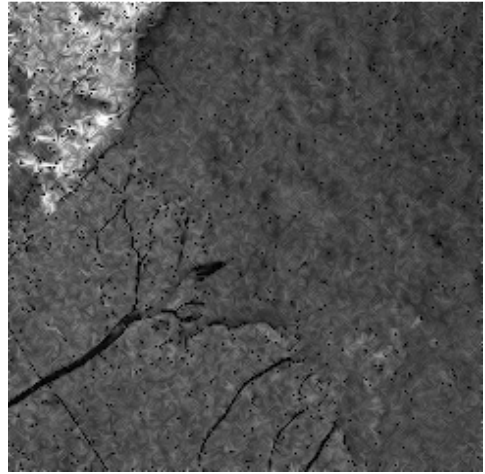

(b) VisuShrink

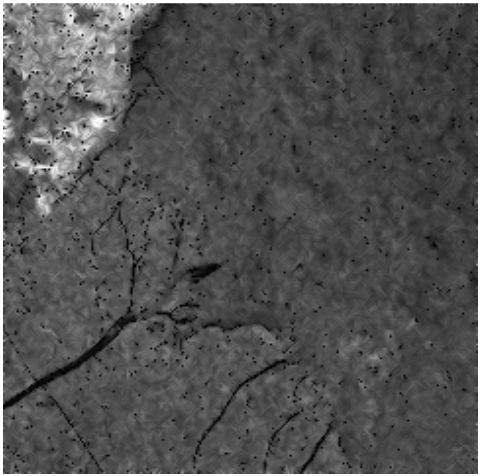

(c) BayesShrink

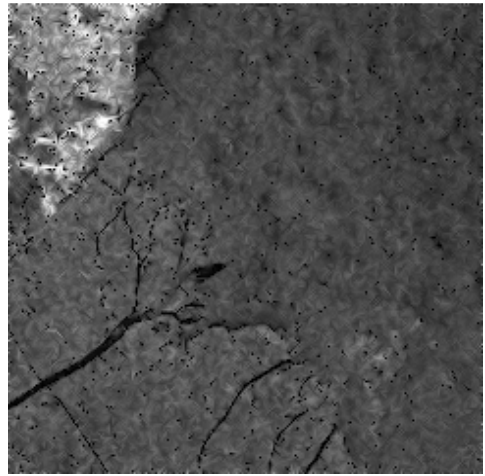

(d) OracleShrink

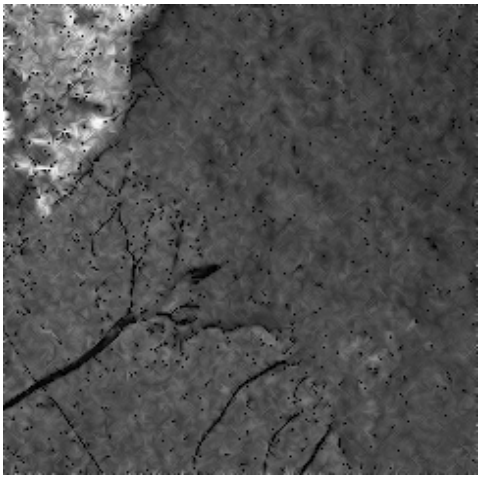

(e) SUREShrink

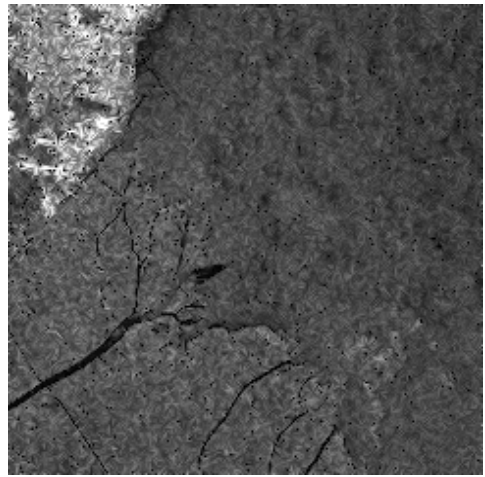

(f) FuzzyThresh

Fig. 16: Original and filtered images.

$$\Gamma(\bullet) = (1+th[\bullet])/2 \qquad (12)$$

where *th*[•] means hyperbolic tangent, while for the output layer the activation functions were the thresholding operator. I consider that it is important to mention that, morphology-based nonlinear filters [55], [56] were used without better results that the worst in the simulated ones.

Finally, all filters were implemented in MATLAB® (Mathworks, Natick, MA) on a PC with an Athlon (2.4 GHz) processor.

## VIII. CONCLUSION

In this paper I have developed a FuzzyThresh algorithmic version based techniques for removing multipli-cative noise in SAR imagery. The simulations show that the FuzzyThresh have better performance than the most commonly used filters for SAR imagery (for the studied benchmark parameters) which include statistical filters (with 3-by-3, 5-by-5, and 7-by-7 kernel windows), VisuShrink (ST, HT and SST) with Daubechies wavelet basis 15 and 1 level of decomposition, SureShrink, Oracle-Shrink, BayesShrink, OracleThresh and a version of TNN. This observation has directed me to formulate a new adaptive edge-preserving application of the FuzzyThresh tailored to speckle contaminated imagery.

The FuzzyThresh exploits the local coefficient of variations in reducing speckle. The performance figures obtained by means of computer simulations reveal that the FuzzyThresh algorithm provides superior performance in comparison to the above mentioned filters in terms of smoothing uniform regions and preserving edges and features. The effectiveness of the technique encourages the possibility of using the approach in a number of ultrasound and radar applications. Besides, the method is computationally efficient (although but high regarding its competitors) and can significantly reduce the speckle while preserving the resolution of the original image. Considerably increa-sed deflection ratio strongly indicates improvement in detection performance.

The optimal parameters of this applied algorithm to get better speckle reduction depends on image characteristics, and statistics of noise. In facts, the novel demonstrated to be efficient to remove multiplied noise, and all uncle of noise in the undecimated wavelet domain. On the other hand, cleaner images suggest potential improvements for classifycation and recognition. Besides, the natural extension of this work is in medical applications, as well as in microarrays denoising.